\documentclass[10pt, conference, compsocconf]{IEEEtran}
\usepackage{url}
\usepackage{tabularx}

\newtheorem{thm}{Theorem}
\newtheorem{defin}[thm]{Definition}

\usepackage{lipsum}

\newcommand\blfootnote[1]{%
  \begingroup
  \renewcommand\thefootnote{}\footnote{#1}%
  \addtocounter{footnote}{-1}%
  \endgroup
}

\ifCLASSINFOpdf
 \usepackage[pdftex]{graphicx}
\else
\fi
\usepackage{color}
\usepackage{multirow}
\usepackage[cmex10]{amsmath}

\usepackage{amssymb}
\usepackage{lmodern}

\usepackage{algpseudocode}
\usepackage{algorithmicx}
\usepackage{algorithm}

\usepackage{array}

\usepackage[skip=0pt]{caption}
\usepackage{subcaption}
\usepackage{float}

\begin{document}
%
\title{Terminology-based Text Embedding for Computing Document Similarities on Technical Content}


\author{\IEEEauthorblockN{Hamid Mirisaee}
\IEEEauthorblockA{Skopai\\Grenoble, France\\
Hamid.Mirisaee@skopai.com}
\and
\IEEEauthorblockN{Eric Gaussier}
\IEEEauthorblockA{Univ. Grenoble Alps/CNRS\\Grenoble, France\\
Eric.Gaussier@imag.fr}
\and
\IEEEauthorblockN{Cedric Lagnier}
\IEEEauthorblockA{Skopai\\Grenoble, France\\
Cedric.Lagnier@skopai.com}
\and
\IEEEauthorblockN{Agnes Guerraz}
\IEEEauthorblockA{Skopai\\Grenoble, France\\
Agnes.Guerraz@skopai.com}}


\maketitle

\begin{abstract}
We propose in this paper a new, hybrid document embedding approach in order to address the problem of document similarities with respect to the technical content. To do so, we employ a state-of-the-art graph techniques to first extract the keyphrases (composite keywords) of documents and, then, use them to score the sentences. Using the ranked sentences, we propose two approaches to embed documents and show their performances with respect to two baselines. With domain expert annotations, we illustrate that the proposed methods can find more relevant documents and outperform the baselines up to 27\% in terms of NDCG.
\end{abstract}

\section{Introduction}\label{sec:intro}
\blfootnote{This article has been published in the Proceedings of the TALN-RECITAL 2019 conference. The original manuscript is available on the ACL Anthology website: \url{http://www.aclweb.org/}}
Many Machine Learning (ML) applications require the calculation of similarities between instances of the task under consideration. Those instances could be of any nature such as numerical and/or categorical time series \cite{aghabozorgi2015time}, gene expression \cite{wang2017visualization} or textual documents \cite{nikolentzos2017shortest, song2015unsupervised}. 

Perhaps the application of similarity assessment between objects that is vastly used on an everyday basis is the one used in search engines. As the most evident example, when a user is looking, via a given query and a search engine such as Google, for relevant documents, there is basically a matching process happening behind. In such application, the user provides a query, as a set of keywords, and looks for the documents which best correspond to those keywords. This mainly leads to a ranking problem where the goal is to rank all the available documents (webpages in case of a web search engine) and provide the user with the most relevant documents. Session search \cite{guan2012effective} is an interesting challenge of such applications where the user reformulates the query based on the results returned by the system until the desired documents are found.

As one can notice, the principal notion of the above-mentioned task is the similarity between instances. For example, a very common similarity measure between textual documents is the tf-idf which is based on two notions: the frequency of the terms appearing in the document, \emph{tf}, and the importance of those terms through the entire set of documents, \emph{i.e.} inverse document frequency, \emph{idf}. Many studies still use tf-idf to perform text-related task as it can, for many tasks, properly project the textual data into the numerical space such that the content is reflected accordingly. One can then use the tf-idf vectors of documents to measure the similarity between them, via cosine for instance. The tf-idf representation is widely used in different applications such as document clustering \cite{bafna2016document} and topic modeling \cite{zhu2013building}.

As one can see, tf-idf embeds the document by operating at the word-level, \emph{i.e.} it takes the tf-idf score of each word and represent the document as a (sparse) vector of size of the entire vocabulary. There is, however, another recent word-level operated technique, namely word2vec \cite{mikolov2013efficient}, which also aims at representing words. Nevertheless, there are two main differences between these two approaches. Firstly, tf-idf assigns a scalar to each word while word2vec represents each word by a vector. Secondly, and more importantly, word2vec embeds each word using its context, \emph{i.e.} its surrounding words. That being said, it is able to capture the semantical context of words. The latter has shown very good results in many applications such as text clustering and text classification \cite{wang2015semantic, lilleberg2015support}.

As mentioned previously, calculating document similarity and, consequently, finding similar documents is at the core of many ML tasks. Sometimes, however, focusing on the entire document may not lead to capturing desired similar documents as not all parts of a document have the same importance level. For instance, if a document describes a novel device for people suffering from diabetes, then taking the entire document may not necessarily result in the similar documents talking about the same particular issue, but rather about the medical domain in general. Note that this issue is different from data cleaning, and should be rather considered as data selection/weighting for the task of document embedding.

In this paper, we investigate the above-mentioned problem and propose a technique for document embedding w.r.t. technical content of the documents. We show that the proposed method is able to find better similar documents once the technical content is concerned which, to the best of our knowledge, is the first research targeting this objective. To do so, we propose to first capture the keyphrases, \emph{i.e.} composite keywords, of the document and, then, rank the sentences based on the keyphrases they contain. To detect keyphrases, we use the state-of-the-art technique explained in \cite{rousseau2015main} which is based on the k-core concept of graph-of-words. We then use the two previously-mentioned embedding techniques to represent the text and show that the proposed, hybrid method can efficiently capture relevant documents based on the technical content.

The reminder of this paper is organized as follows: Section~\ref{sec:related} overviews the most related studies. Section~\ref{sec:framework} first details the graph representation of documents and the k-core concept, before providing the framework of embedding documents using the sentence ranking obtained via the k-core approach. Then, in Section~\ref{sec:exp}, we describe the experimental settings and the baselines as well as the collected dataset. We then report our results in the same section. Finally, Section~\ref{sec:conclusion} concludes the paper.

\section{Related work}\label{sec:related}
Over the past few decades, a large body of studies has been considering the relatedness of documents for a wide range of tasks such as text categorization or document classification. Assessing the similarities between documents is at the core of many machine learning applications such as information retrieval, recommendation systems and text generation. In this section, we consider the most relevant studies with respect to the topic under investigation and position our work in regard to them.

Classically, the similarity between two documents has been measured by the cosine similarity between their tf-idf vectors, see \cite{trstenjak2014knn} and \cite{schultz1999topic} for example. Alternatively, studies like \cite{zhang2011comparative} considered other techniques such as LSI and multi-word methods and investigated their performance in different tasks. Although tf-idf can still capture many characteristics of a document and boost the performance of many tasks such as topic modeling \cite{hong2010empirical, mehrotra2013improving}, they still fail to detect the entire context of a document such as the semantic relation between words \cite{tapaswi2016movieqa}.

The paper of Mikolov et al. \cite{mikolov2013efficient} introduced an entire new idea of textual representation where a document is analyzed at the word level and each word is represented based on its context, \emph{i.e.} it carries semantic features of the document. The context of the word is defined by the co-occurring words. The model is simply a neural network where the vectorial embedding of a word is determined by those of its surrounding words. Many variations of this approach have been later studied to, for instance, operate directly at the document level \cite{lau2016empirical, mikolov2013distributed}. In addition, the combination of tf-idf and word2vec has been widely used and shown to bring significant improvement in many applications \cite{lilleberg2015support, acosta2017sentiment}. As it will be illustrated in Section~\ref{sec:exp}, we will make use of the combination of these two techniques to conduct a part of our experiments, and will show that they work very efficiently in capturing the technical side of documents.
 
Measuring the similarity between documents has been widely studied in the literature through different approaches. For instance, \cite{cooper2002detecting} used a phrase recognition approach to detect similar documents, and  \cite{pereira2003syntactic} employed radix tree to calculate similarities between web documents. In a different approach, \cite{paul2016efficient} introduces a graph-based method to exploit the hierarchical relations in order to efficiently calculate the similarity between documents. Similarly, \cite{wang2015knowsim} proposes to represent documents as typed heterogeneous information networks and, following the notion of graphs, it computes the distance between documents for the task of document clustering.
 
Perhaps the most related studies to ours are \cite{brants2002finding} and \cite{cooper2002novel}. In \cite{brants2002finding}, the authors propose to use a word-based method to capture similar documents via PLSA. The document matching in this particular work is done based on the words appearing in the document. More precisely, the similarity between two documents is defined as linear combination of the cosine between the tf-idf vectors and the PLSA-based representations of the words. Although the objective of this paper is similar to our work in that they also rely on words-basis scoring, two main differences distinguish that research and ours. Firstly, we focus on keyphrases and pay more attention on the parts of the text where those keyphrases are used, while \cite{brants2002finding} relies on the entire content of the document. Secondly, the present study uses the context-based techniques, such as word2vec to particularly target the semantic aspects of a document. In other words, if two documents describe the similar subjects with different vocabulary, the purely tf-idf based or topic modeling based approaches fail to see their similarities, while context-based techniques are able to capture that. In a similar manner, \cite{cooper2002novel} defines two documents to be similar if they have the same pieces of text such as sentences or paragraph, which has similar limitations as mentioned before. This current study, to the best of our knowledge, is the first one to propose technical content based similarity between documents using their semantic terminology.
\section{Framework}\label{sec:framework}
Graph-based techniques have been widely used in representing textual documents, where, in general, meaningful linguistic units of the text, such as paragraphs \cite{balinsky2011automatic}, sentences \cite{mihalcea2004textrank} or words \cite{schenker2003graph}, construct the nodes and the relation between them defines the edges. This relation could be of different natures depending on the task in mind. statistical (such as co-occurrence) and syntactic (such as noun-adjective relation) are two widely used kind of relations between nodes of the graph.

Following this line of thought, \cite{rousseau2015main} investigated the task of keyword extraction by representing a document as a \emph{graph-of-words} and retrieving the \emph{main core} of the graph. In the following, we briefly explain how this task is accomplished and, then, use the results of this technique to rank the sentences and eventually embed a document.

\subsection{Graph-of-words}\label{subsec:gow}
In \cite{rousseau2015main}, the authors propose to represent a document as a graph where nodes are terms of the document. Two nodes are then connected via an edge if they co-occur within a fixed-size window. More formally:
\begin{defin}
The graph-of-words of document $d$ is defined as  $\mathcal{G} = (\mathcal{V}, \mathcal{E})$ where $\mathcal{V}$ is the set of nodes that represents the terms of $d$ and $\mathcal{E}$ is the set of edges which indicates the co-occurrence of the terms within a fixed-size sliding window of size $n$.
\end{defin}
Note that $\mathcal{E}$ can be weighted or unweighted. In the weighted setting, the weight of $e_{ij}$ is the number of times that $w_i\in\mathcal{V}$ and  $w_j\in\mathcal{V}$ co-occur in the sliding window within a document. One can also consider a directed (weighted/unweighted) version of where the order of words in the document determines the direction of edges. In other words, if $w_1$ precedes $w_2$ in the document within the sliding window then the edge $w_1\rightarrow w_2$ is added to $\mathcal{E}$.

With such representation of the document, \cite{rousseau2015main} proposes a technique which focuses more on cohesiveness of the nodes, rather than classic methods, such as \cite{litvak2008graph} and \cite{mihalcea2004textrank}, which rely on the notion of centrality. To do that, the \emph{k-core} approach \cite{seidman1983network} has been employed:
\begin{defin}
$\mathcal{H}_k = (\mathcal{V}^\prime, \mathcal{E}^\prime)$ is called a k-core or a core of order $k$ of $\mathcal{G} = (\mathcal{V}, \mathcal{E})$ iff $\mathcal{E}^\prime\subset  \mathcal{E}$, $\mathcal{V}^\prime\subset  \mathcal{V}$ and $\forall v\in \mathcal{V}^\prime$, $Deg(v)\ge k$, and  $\mathcal{H}_k$ is the maximal graph with such property. The core of maximum order is then called the main core of $\mathcal{G}$.
\end{defin}
The intuition behind using the $k$-core approach is to not only focus on the central nodes of the graph, but also pay attention to how connected the neighbors of the the node are, which is known as the cohesion of a graph. Following this notion, \cite{rousseau2015main} proposes to use the $k$-core approach for keyword extraction.

Basically, the idea is to, starting from the graph-of-words, calculate the $k$-core of the graph and then take all the nodes of the main core, \emph{i.e.} the core with the maximum order, as keywords. Via extensive experiments, the authors show that their approach outperforms the traditional methods such as HITS and PageRank \cite{litvak2008graph, mihalcea2004textrank}. Additionally, unlike other techniques that need the number of keywords to extract, the size of the main core basically handles this issue. On top of that, this technique can effectively be used to detect keyphrases, \emph{i.e.} composite keywords\cite{rousseauPhD}. Once again, in such context, each term of the text is represented as a node in the graph and two connected nodes can potentially construct a keyphrase.

For example, in the sentence "our platform is based advanced artificial intelligence techniques", both "artificial" and "intelligence" can be considered as keywords, and "artificial intelligence" can construct a keyphrase. Note that if the size of sliding window explained previously is larger than 2, then keyphrases like "artificial techniques" and "advanced intelligence" can also be extracted from the graph as keyphrases.

With that in mind, one can simply use the nodes of the main core to construct the keyphrases using different approaches, by for instance taking the words corresponding to the connected nodes. We use a slight modification of the k-core method in order to, first, extract the keyphrases of size 2, \emph{i.e.} combination of two and only two terms, and, then, rank the sentences. Using those ranked sentences, we propose a technique to embed the document such that it encodes the main technical content of the document. The proposed approach is detailed in the following.

\subsection{\texttt{TDE}: Terminology-based Document Embedding}
As explained above, the k-core, graph-based methods can efficiently be used to extract keywords or keyphrases, where keyphrases often better reflect the semantics of the  document as they tend to reduce the noise significantly. Knowing that the objective of this study is to eventually embed a document w.r.t. its technical content through the keyphrases, we first establish a link between the embedded information in the graph-of-words and the resultant keyphrases. We then score the sentences based on the keyphrases they contain and, finally, calculate the embedding of the document via its scored sentences. In other words, the embedding of a document is a weighted average of the embedding of its sentences, where the weights are derived from the graph-of-words which has been explained above.

Let $\mathcal{C}=\{c_1,\cdots, c_k\}$ be the set of all cores of the graph-of-words of document $d$ where $k$ is the maximal order of the graph and $c_k$ is the maximal core. Also, let $\mathcal{T}_{c_i}$ be the set of all keyphrases  appearing in core $c_i\in\mathcal{C}$. More formally, $\mathcal{T}_{c_i} = \{(t,t^\prime)|t\in c_i \wedge t^\prime\in c_i\}$.
Then, the embedding of the document $d$, denoted as $\vec{d}$, can be calculated as the weighted average of the embedding of its sentences:
\begin{equation}\label{eq:vec_d}
\vec{d} = \frac{1}{\sum_{s\in S} \Gamma(s)}\sum_{s\in S} \vec{s} \times \Gamma(s)
\end{equation}
where $S$ is the set of sentences of $d$, $\vec{s}$ is the embedding of the sentence $s$ and $\Gamma(s)$ is the score of sentence $s$.
Eq.~(\ref{eq:vec_d}) is actually a weighted average of the embeddings of all the sentences of the document. Here, the idea is to use the keyphrases of each sentence to determine its score (weight). We propose the following for calculating the score of each sentence $s$ of the document:
\begin{equation}\label{eq:all_cores}
\Gamma(s) = \sum_{i=1}^{i=k} \sum_{\substack{(t,t^\prime)\in \mathcal{T}_{c_i} \\ (t,t^\prime)\in s}} \phi\big((t,t^\prime)\big)
\end{equation}
where the function $\phi(.)$ returns a score (weight) for each keyphrase $(t,t^\prime)$. This is mainly where the graph information is taken into account. The score of the keyphrase $(t,t^\prime)$ can be calculated using its properties found in the k-core setting, \emph{i.e.} the degree of the edge connecting  $t$ and $t^\prime$, as well as the core where those two nodes (terms) appear:
\begin{equation}\label{eq:F_func}
\phi\big((t,t^\prime)\in \mathcal{T}_{c_i}\big) = Deg(t, t^\prime)\times F(c_i)
\end{equation}
where $F(.)$ is a function that assigns a weight to each core such that cores get monotonically decreasing weights. Obviously the main core has the highest weight.
In our experiments, we use the rational function:
\begin{equation}\label{eq:ratio}
F(c_i)=(k-i+1)^{-1}
\end{equation}
where $k$ is the maximum order of the cores.

We implemented the graph-words as an undirected weighted graph where the number of co-occurrences of two words determines the weight of the edge linking them. One should note that the directed version has also been investigated and, according to our observations, was not as good as the undirected one. For the embedding of the sentences (Eq.~(\ref{eq:vec_d})) we propose two different techniques which are further explained in the next Section. Finally, it should be noted that unlike \cite{rousseau2015main}, we use all the cores of the graph as it is shown in Eq.~(\ref{eq:all_cores}) where the more focus is still on the most important cores as expressed in Eq.~(\ref{eq:ratio}).

Algorithm~\ref{algo:TDE} illustrates the procedure of \texttt{TDE} regardless of the embedding chosen for the sentences of the document. Needless to say, in practice the algorithm can be accelerated if the loops at lines \ref{L2} and \ref{L3} are done via memory operations. This can be simply done by precomputing and storing all the possible keyphrases (all adjacent nodes of the graph) and their scores, in a dictionary-like data structure for instance.

\begin{algorithm}[t]
	\caption{Terminology-based Document Embedding}
	\begin{algorithmic}[1]
	\Require Set $S$ containing all sentences of document $d$, $\mathcal{T}_{c_i}\,(1\leq i\leq k)$: keyphrases of each core
	\Ensure $\vec{d}$: the embedding of $d$
	\State $w=0$
	\State $\vec{d}=\vec{0}$
	\ForAll{$s\in S$}
		\State $w_s=0$
		\For{$i=1$ to $k$}	 \label{L2}
			\ForAll{$(t,t^\prime)\in \mathcal{T}_{c_i}$} \label{L3}
				\If{$(t,t^\prime)\in s$}
					\State $w_s = w_s + \frac{Deg(t, t^\prime)}{i}$ \quad\quad \textbackslash\textbackslash\; Eqs.~(\ref{eq:F_func})-(\ref{eq:ratio})
				\EndIf
			\EndFor
		\EndFor
		\State $\vec{d} = \vec{d} + (w_s\times\vec{s})$ \quad\quad \textbackslash\textbackslash\;$\vec{s}$ is the embedding of $s$
		\State $w = w + w_s$
	\EndFor	
	\State $\vec{d} = \frac{\vec{d}}{w}$
	\State RETURN $\vec{d}$
  \end{algorithmic}
  \label{algo:TDE}
\end{algorithm}
\section{Experiments}\label{sec:exp}
\subsection{Baselines}\label{subsec:baselines}
To evaluate the validity of the proposed technique, we compare the method explained in Section~\ref{sec:framework} with two document embedding baselines.
As mentioned in Section~\ref{sec:related}, many studies investigate the problem of document embedding as an end-to-end problem where the document is embedded all at once, \emph{i.e.} the embedding is learned at the document level. Accordingly, we chose, as our first baseline, the document embedding method investigated in \cite{lau2016empirical} which is a robust improvement over the original \emph{doc2vec} \cite{le2014distributed}. This baseline will be denoted as \texttt{D2V}. Although \texttt{D2V} is not designed particularly for the task that we are investigating, \emph{i.e.} finding similar documents w.r.t. technical content, we still keep it as a baseline as it is one of the popular techniques to address the document embedding problem.

Following the ideas presented in \cite{wieting2016charagram} and \cite{arora2016latent}, one can compute the document embedding by averaging the embedding of the words appearing in the document. The word embeddings can be learned via classical methods such word2vec explained in \cite{mikolov2013efficient}. According to our experiments, tf-idf weighted average performs significantly better than simple averaging and, consequently, we chose to use it as the second baseline. This baseline is referred to as  \texttt{TWA} (Tf-idf Weighted Average).

These two baselines are then compared to the graph-based method proposed in this paper and detailed in Section~\ref{sec:framework}. As mentioned in that section, a document is represented as a weighted average of the embeddings of the sentences appearing in it, where the weights are the score of sentences which, in turn, are calculated using the graph-based representation of the document (see Eqs.~(\ref{eq:vec_d})-(\ref{eq:ratio})). To be consistent with the abbreviation of the previous section, this technique will be denoted as \texttt{TDE}.

In \texttt{TDE}, the embedding of a sentence can be calculated in different ways. Here, we propose two ways of doing that: the first one consists of using the models which are trained to directly produce the sentence embedding; and the second one is to use the embeddings of the words forming the sentence. For the former, the state-of-the-art sentence embedding technique, described in \cite{pagliardini2018unsupervised} is employed. For the latter, we perform a tf-idf weighted average of the words constructing the sentence to calculate the embedding of the sentence. One should note that as in a sentence it is rarely the case that we have repetition of words, the calculation is almost an idf weighted averaging. To avoid an ambiguity, the first approach is denoted as \texttt{TDE}$_{\texttt{s2v}}$ (for sent2vec) and the second as \texttt{TDE}$_{\texttt{iw}}$ (for idf weighted).
In the following, we explain our dataset from which the embeddings of words, sentences and documents are learned.
\subsection{Dataset, embedding models and preprocessing }\label{dataset_and_more}
We crawled websites of around 68K startups (all around the world, with no constraint on the  domain of business) with a total number of 3.4M webpages. After filtering those without sufficient textual information or non-English ones, we ended up with around 43K startups and 2.8M pages. We only used the English sentences of the pages for training our models and performing the evaluations.

To train the word2vec, we used \emph{gensim}\footnote{\url{https://radimrehurek.com/gensim/}} on the sentences extracted from the 2.8M pages mentioned above, with a minimum count of words and window size equal to five. The number of sentences extracted reached to 950K. We used the authors' implementation\footnote{\url{https://github.com/epfml/sent2vec}} of \cite{pagliardini2018unsupervised} to train the sent2vec model.

Furthermore, keeping in mind that the objective of this study is to investigate document similarities based on the technical content, not all parts of a document are of interest. Knowing that each document in our dataset is the combination of all the textual content of a startup's website, many parts of that can possibly be considered as noise. For instance, the pages describing privacy policies or legal information must be ignored before performing any document embedding process. As a result, we used multiple classifiers, trained using thousands of pages, to filter out such contents.
Accordingly, as the first preprocessing step, we trained three separate SVM classifiers to filter out the pages with privacy, legal information or cookies information.

\subsection{Evaluation}
To fairly evaluate the performance of the above-mentioned baselines, \texttt{TWA} and \texttt{D2V}, with that of the proposed ones, \texttt{TDE}$_{\texttt{s2v}}$ and \texttt{TDE}$_{\texttt{iw}}$, we adapted the following strategy: we selected a set of 100 documents (startups)  from four different domains: medical, agriculture, energy and biology. Evidently, the texts of these documents have not been used for training the above-mentioned models.

For each test case and for each method, we extracted the 5 most similar startups (using cosine similarity between the indexed embedding of each method and the entire dataset). The results are then combined, shuffled and assigned to the corresponding domain expert for evaluation. The experts then give a score to each of the startups from the score set \{1, 2, 3, 4, 5\} where 1 denotes the least relevant and 5 denotes the most relevant. Note that the experts have been clearly informed that the combined list is not in any way ordered/ranked and the outputs of a test case should be evaluated independently. Obviously, the results do not necessarily contain all the possible scores.

To assess the performance of the proposed approach, we use NDCG (Normalized Discounted Cumulative Gain) which perfectly matches our experimental settings. We report here two values, namely NDCG@1 and NDCG@5, to investigate the behavior of each technique. In the following, we discuss the results of our experiments and illustrate the efficiency of the proposed methods.

\subsection{Results}
Table~\ref{tab:exp} shows the results of our experiments on all four approaches detailed in Section~\ref{subsec:baselines} using NDCG@1 and NDCG@5, compared via t-test at 5\% for the significance test. As one can notice, the performance of \texttt{D2V} falls way below other methods with both NDCG@1 and NDCG@5. That could be explained by the fact that \texttt{D2V} needs long documents to be trained properly, and since our documents are only English sentences of only a part of a website (see the filtering procedures explained in Section~\ref{dataset_and_more}), they could be as short as few sentences. As a consequence, \texttt{D2V} has difficulties to project the context into the numerical space. As a result, we will not investigate this technique in our discussion below.

\begin{table}[t]
\centering
\begin{tabular}{lcccc}
       & \texttt{D2V} &  \texttt{TWA} & \texttt{TDE}$_{\texttt{iw}}$ & \texttt{TDE}$_{\texttt{s2v}}$ \\ \hline\hline
NDCG@1 & 0.26   & 0.54   & 0.63       & \textbf{0.69}      \\ \hline
NDCG@5 & 0.24   & 0.60   & 0.60       & \textbf{0.65}
\end{tabular}
\caption{NDCG@1 and NDCG@5 on all four methods. The best approach is shown in bold.}
\label{tab:exp}
\end{table}

The second baseline, \emph{i.e.} \texttt{TWA}, performs well w.r.t. the proposed methods; as one can observe, it achieves the same NDCG@5 compared to  \texttt{TDE}$_{\texttt{iw}}$. However, if sent2vec is used to embed the sentences, then one can outperform \texttt{TWA} by \textbf{8\%} in terms of NDCG@5 via  \texttt{TDE}$_{\texttt{s2v}}$

When it comes to NDCG@1, both graph-based variants outperform the baselines significantly: \texttt{TDE}$_{\texttt{iw}}$ is around \textbf{16\%} better that \texttt{TWA} and \texttt{TDE}$_{\texttt{s2v}}$ has a better NDCG@1 value by \textbf{27\%} w.r.t. \texttt{TWA}.

These experiments show that the proposed technique, be it with the tf-idf weighting or with the sent2vec method, outperform the baselines. In other words, they are able to capture better similar documents when the technical content is of interest. Additionally, according to the results reported on NDCG@1, the proposed methods can find way better results when finding the most relevant document is the task in mind. This could be particularly interesting as in many IR tasks, such as search engines, the first retrieved document plays a very important role in the further processing steps \cite{joachims2005accurately}.

\section{Conclusion}\label{sec:conclusion}
In this paper, we have studied the problem of finding similar documents with respect to technical content. To do that, we proposed a hybrid approach which employs the graph representation techniques and sentence embedding. Using the graph-of-words representation of a document and computing the cores of the graph, we first extract the keyphrases of the document of all cores of the graph. We then use this information to assign a score to each sentence; once the scores are calculated, we use the embedding of each sentence and its score to calculate the embedding of the document. We proposed two approaches to embed sentence: (\emph{i}) sent2vec, a state-of-the-art technique for sentence embedding, and (\emph{ii}) a tf-idf weighted average of the words appearing in the sentence. We then compared those to two baselines, \texttt{doc2vec}, an existing method in the state-of-the-art to embed documents, and a tf-idf weighted average of the words appearing in the document.

As dataset, we used 2.8M webpages of 43K startups that we crawled from the web, where we considered the combination of all webpages of a startup as a document. To evaluate, we asked human experts to score the output of all techniques. Using the NDCG metric, we illustrated that the proposed metric can outperform the baselines up to 27\% and, as a result, can provide better similar documents when technical content is concerned.

\section*{Acknowledgments}
The authors would like to thank the domain experts of Skopai, Adeline Tarantini, Olivier Berengario, Elie Gehin and Guillaume Emery who spend a considerable amount of time to carefully evaluate the results.

\bibliographystyle{IEEEtran}

\bibliography{main}

\end{document}